%% file: acl.tex
\documentclass[11pt]{article}

\usepackage{acl}

\usepackage{times}
\usepackage{latexsym}

\usepackage[T1]{fontenc}

\usepackage[utf8]{inputenc}

\usepackage{microtype}

\usepackage{url}
\usepackage{array}
\usepackage{amsmath}
\usepackage{amssymb}
\usepackage{caption}
\usepackage{subcaption}
\usepackage{booktabs}
\usepackage{multirow}
\usepackage{graphicx}
\usepackage{dsfont}
\usepackage{balance}
\usepackage{enumitem}
\usepackage{xcolor,colortbl}
\usepackage{makecell}
\usepackage{bbm}
\usepackage{soul}
\usepackage{blindtext}
\usepackage[linesnumbered,ruled]{algorithm2e}
\usepackage{pifont}

\newcolumntype{P}[1]{>{\centering\arraybackslash}p{#1}}
\title{Striking a Balance: Alleviating Inconsistency in Pre-trained Models for Symmetric Classification Tasks}

\author{Ashutosh Kumar \\
  Indian Institute of Science, India \\
  \texttt{ashutosh@iisc.ac.in} \\\And
  Aditya Joshi \\
  SEEK, Australia  \\
  \texttt{aditya.m.joshi@gmail.com},\\
  \texttt{ajoshi@seek.com.au}\\
}
  
\begin{document}
\maketitle
\input{defs}
\begin{abstract}
While fine-tuning pre-trained models for downstream classification is the conventional paradigm in NLP, often task-specific nuances may not get captured in the resultant models. Specifically, for tasks that take two inputs and require the output to be invariant of the order of the inputs, inconsistency is often observed in the predicted labels or confidence scores.
We highlight this model shortcoming and apply a consistency loss function to alleviate inconsistency in symmetric classification. Our results show an improved consistency in predictions for three paraphrase detection datasets without a significant drop in the accuracy scores. We examine the classification performance of six datasets (both symmetric and non-symmetric) to showcase the strengths and limitations of our approach.

\end{abstract}

\input{sections/introduction}

\input{sections/relatedwork}
\input{sections/method}

\input{sections/experiment}
\input{sections/results}
\input{sections/conclusion}
\input{sections/ethics}

\bibliographystyle{acl_natbib}
\bibliography{acl}

\input{sections/appendix}
\end{document}

%% file: defs.tex
\newcommand{\refalg}[1]{Algorithm \ref{#1}}
\newcommand{\refeqn}[1]{Equation \ref{#1}}
\newcommand{\reffig}[1]{Figure \ref{#1}}
\newcommand{\reftbl}[1]{Table \ref{#1}}
\newcommand{\refsec}[1]{Section \ref{#1}}

\newcommand{\methoddataset}{\textsc{MTask-Test}}
\newcommand{\methodmain}{\textsc{Framework for Automated Pattern Induction}}
\newcommand{\methodshort}{\textsc{Frapi}}

\newcommand{\reminder}[1]{\textcolor{red}{[[ #1 ]]}\typeout{#1}}
\newcommand{\reminderR}[1]{\textcolor{gray}{[[ #1 ]]}\typeout{#1}}

\newcommand{\remove}[1]{\textcolor{red}{[[REMOVE: #1 ]]}\typeout{#1}}
\newcommand{\add}[1]{\textcolor{red}{#1}\typeout{#1}}

\newcommand{\m}[1]{\mathcal{#1}}

\newcommand{\tensor}{\mathcal{X}}
\newcommand{\Real}{\mathbb{R}}
\newcommand{\tuples}{\mathbb{T}}

\newcolumntype{P}[1]{>{\centering\arraybackslash}p{#1}}

\newcommand\norm[1]{\left\lVert#1\right\rVert}

\newcommand{\note}[1]{\textcolor{blue}{#1}}

\newcommand*{\Scale}[2][4]{\scalebox{#1}{$#2$}}%
\newcommand*{\Resize}[2]{\resizebox{#1}{!}{$#2$}}%

\def\mat#1{\mbox{\bf #1}}

\newcommand{\consistentbert}{%
  \begingroup\normalfont
  \includegraphics[height=11px]{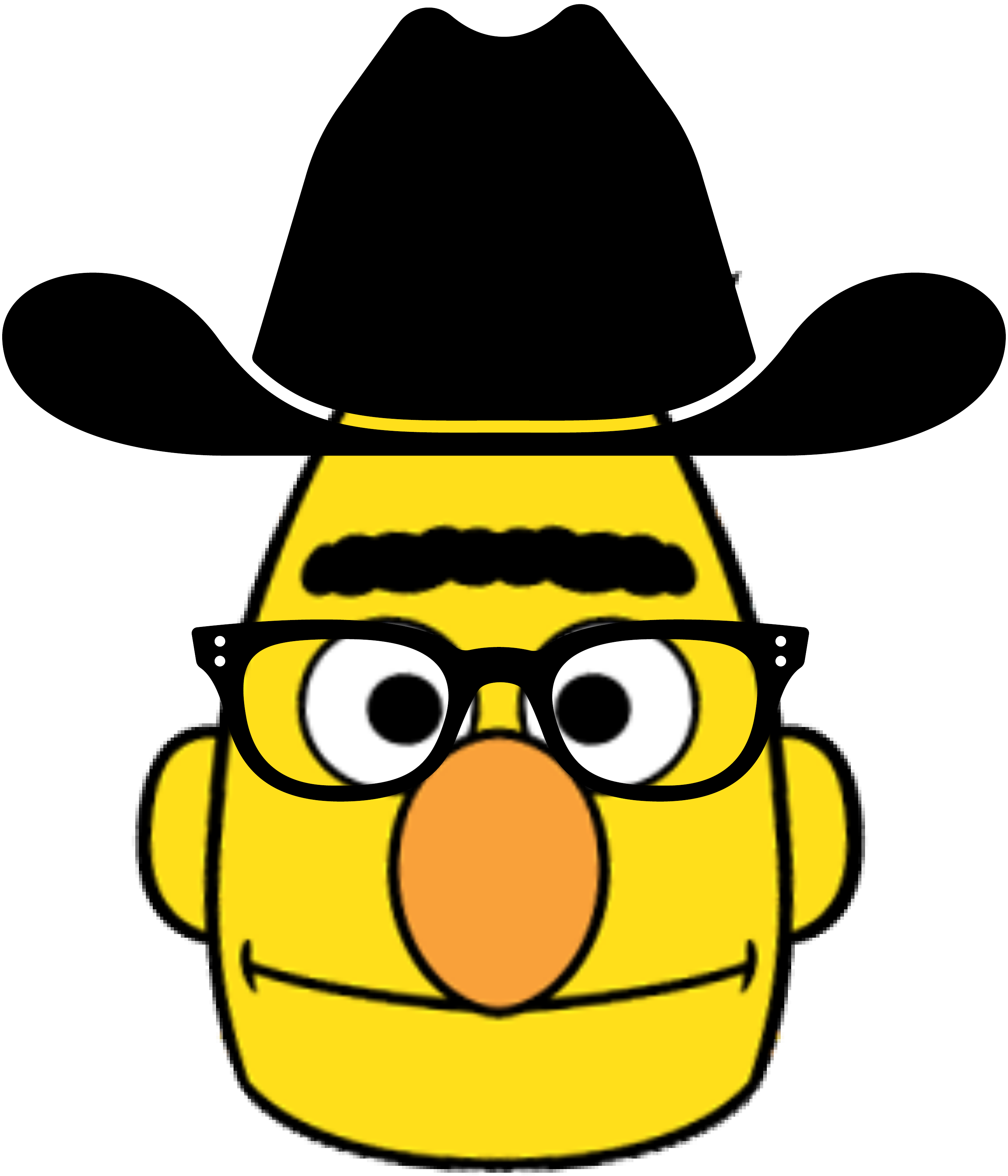}%
  \endgroup
}
\newcommand{\bert}{%
  \begingroup\normalfont
  \includegraphics[height=11px]{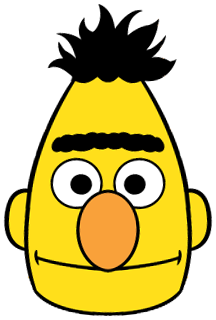}%
  \endgroup
}
\newcommand{\consistentbertsmall}{%
  \begingroup\normalfont
  \includegraphics[height=11px]{images/consistentbert.pdf}%
  \endgroup
}
\newcommand{\bertsmall}{%
  \begingroup\normalfont
  \includegraphics[height=11px]{images/bert_cartoon.png}%
  \endgroup
}

\newcommand{\thumbsup}{%
  \begingroup\normalfont
  \includegraphics[height=11px]{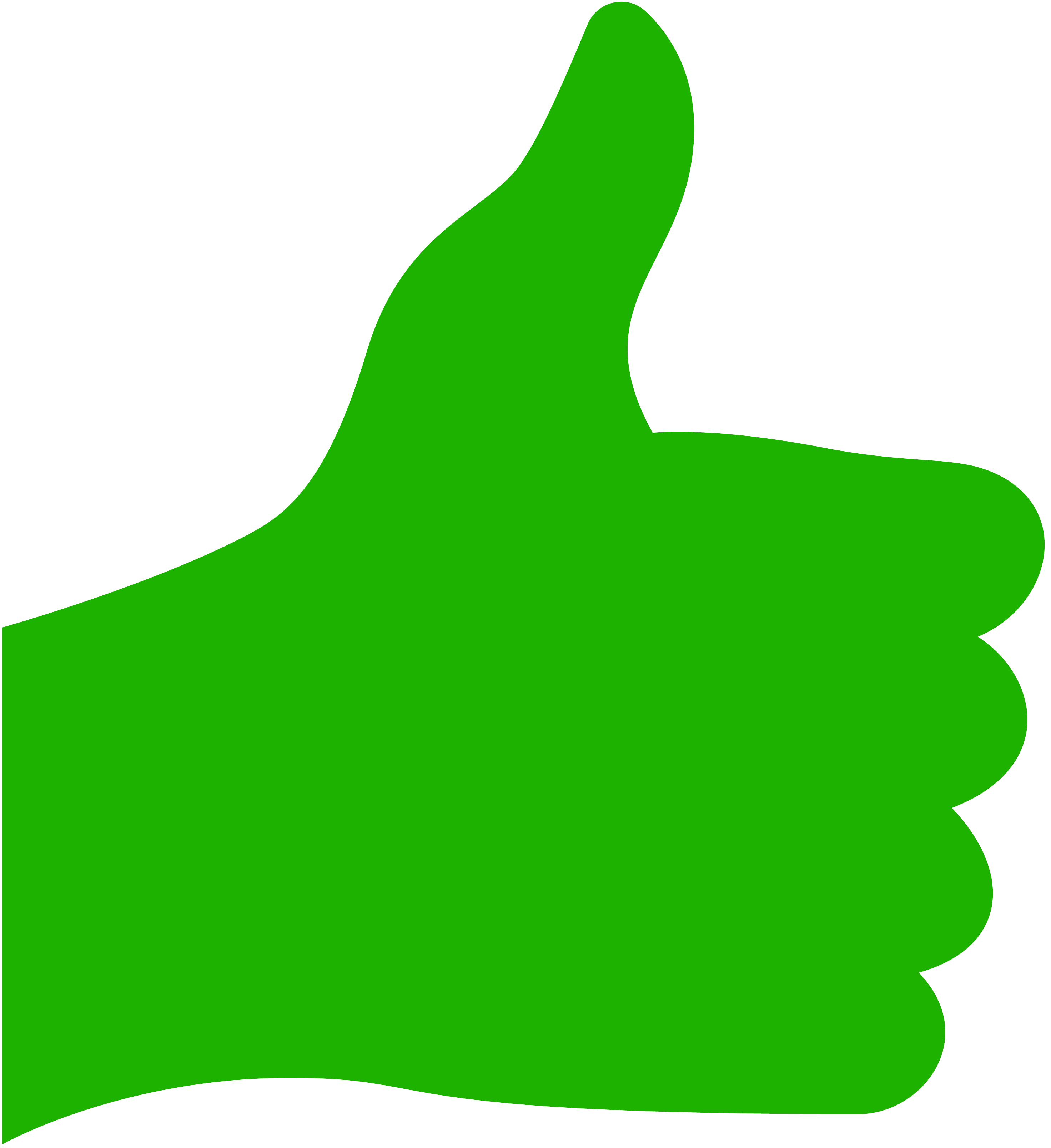}%
  \endgroup
}
\newcommand{\thumbsdown}{%
  \begingroup\normalfont
  \includegraphics[height=11px]{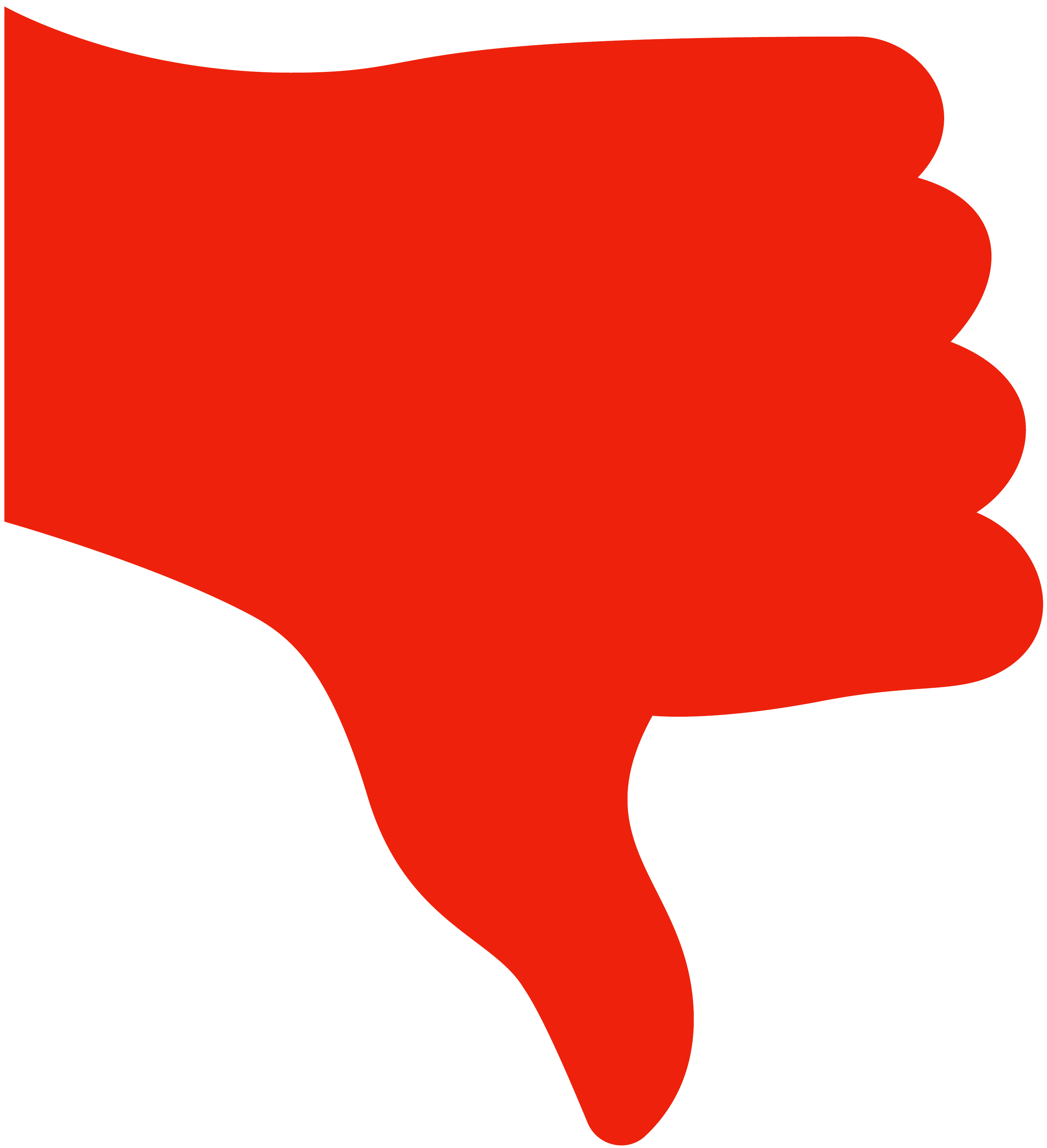}%
  \endgroup
}

%% file: sections/introduction.tex
\section{Introduction}
\label{sec:introduction}
Symmetric classification tasks involve two inputs and require that the model output should be independent of the order in which the two input texts are given. In other words, \textit{the output of the classifier should be the same and the confidence score must not be significantly different}, if the inputs $X$ and $Y$ are instead supplied as $Y$ and $X$. Paraphrase detection, multi-lingual semantic similarity are examples of symmetric classification tasks. 
\begin{figure}[ht]
\centering
\includegraphics[width=0.48\textwidth]{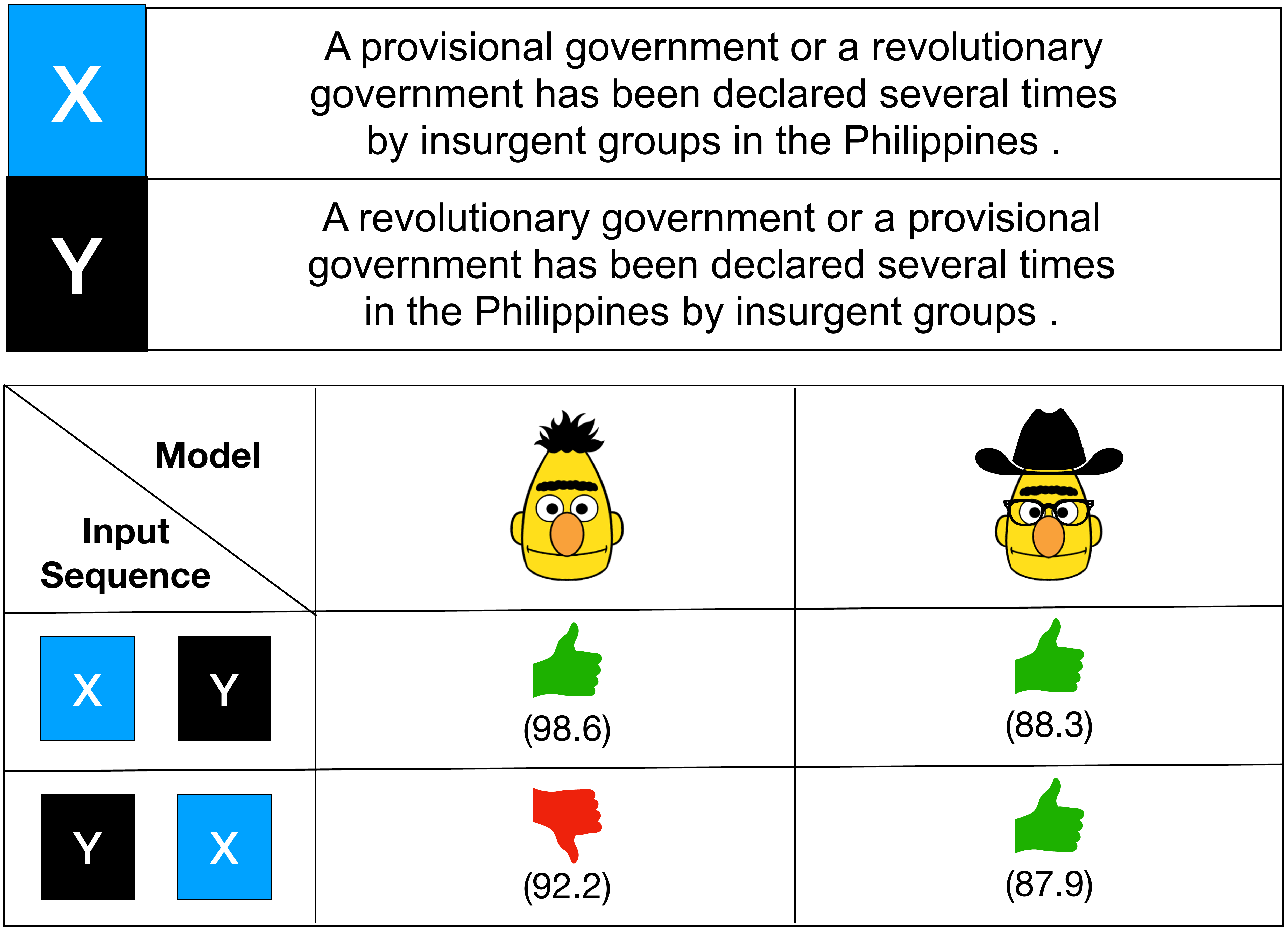}
\caption{\label{fig:firstpage}Impact of reordering an example input pair ($X$ and $Y$) on standard fine-tuned BERT \bert{} and BERT-with-consistency-loss \consistentbert{}. The pair are true paraphrases. \thumbsup{} and \thumbsdown{} denote that the model predicted them to be paraphrases and not-paraphrases, respectively. Confidence scores are reported in brackets. Details in \refsec{sec:introduction}.}
\end{figure}
Although attention-based \cite{bahdanau2016neural, vaswani2017attention} pre-trained language models have led to significant performance gains in multiple text classification tasks, they demonstrate a peculiar erratic behaviour on symmetric classification: inconsistency. An example\footnote{Note that, while this particular example is based on our fine-tuned model, it will change depending on the trained model. The overall argument is valid, nonetheless.} of \textit{inconsistency} for paraphrase detection is shown in ~\reffig{fig:firstpage}. Additional examples can be found in the Appendix (\reftbl{tbl:examples}). To alleviate such an inconsistency for symmetric classification tasks, we propose a simple additional \textit{drop-in} fine-tuning objective, based on either the Kullback-Leibler (KL) or Jensen-Shannon (JS) divergence (or any $f$-divergence \cite{rubenstein2019practical}), to the cross-entropy loss for symmetric tasks. We refer to this as the \textit{consistency loss}.

\noindent The main contributions of this paper are: \\\noindent(a) Highlight inconsistency issues in symmetric classification tasks, \\\noindent(b) Describe a consistency loss function to alleviate inconsistency, and \\\noindent(c) Demonstrate the applicability and limitations of the loss function via qualitative and quantitative analyses on tasks from the GLUE benchmark.  \\
Additionally, to drive future research, we have made the data and code public\footnote{\url{https://github.com/ashutoshml/alleviating-inconsistency}}.

\noindent \textbf{Note:} The problem of inconsistency can be attributed in part to the positional embedding. However, it has been shown that eliminating positional embedding results in a poor performance of the model \cite{wang2020position, wang2021on}.

%% file: sections/relatedwork.tex
\section{Related Work}
\label{sec:relatedwork}
\noindent \textbf{Pre-trained Classification Models} like BERT \cite{devlin2019bert}, and RoBERTa \cite{liu2019roberta}
are typically fine-tuned for classification tasks using a low capacity neural network classifier connected to the pre-trained model on its first token (typically \texttt{[CLS]} token). We demonstrate the inconsistency in the case of symmetric classification tasks for pairs of inputs, depending on the order of inputs. To the best of our knowledge, this is the first work that incorporates task-specific nuances to ensure consistency in symmetric classification.

\noindent \textbf{Consistency Loss} has been used in style transfer tasks to minimize the distance between round-trip generation of candidates for image-to-image translation~\cite{CycleGAN2017} or text style transfer~\cite{huang2020cycle}. In a similar vein, we apply consistency loss (formulated as either the Kullback-Leibler or the Jensen-Shannon divergence loss) to alleviate the inconsistency problem in symmetric tasks.

\noindent \textbf{Embedding-based Semantic Similarity Scores} based on BERT-based models like SBERT \cite{reimers2019sentence, thakur2021augmented} can map surface form realizations to embeddings. Their performance is worse than directly using BERT-style cross-encoder models for tasks such as semantic similarity \cite{thakur2021augmented}. However, the primary aim of such embedding-based scorers is orthogonal and, at best, complementary to the goal of our work since we want to ensure high-performing, consistent classifiers. Similarly, an alternative for symmetric classification is to separately obtain predictions for ($X$, $Y$) and ($Y$, $X$), and then average the confidence scores during test time. But, this is a weakly grounded, heuristic-driven approach. In general, \textit{averaging does not rectify the mistakes made by the model, only masks it.}

%% file: sections/method.tex
\section{Method\label{sec:method}}
\subsection{Problem Description}
\label{subsec:problem}
\textbf{A.} Given a pair of input sentences $(X, Y)$, label $l_{(X,Y)}$, and a pre-trained BERT-based model $\mathcal{M}_{\textsc{Pre}}$, the goal is to output a \textit{reliable model} $\mathcal{M_{\textsc{Rel}}}$ to predict an output label for a new input pair $(X_{test}, Y_{test})$ such that the \textit{inconsistency} between its different ordering is minimized. While we only experiment with semantic similarity (or paraphrasing), the description holds true for other symmetric relations too (such as predicting if two sentences have the same polarity).\\
\textbf{B.} Given a model fine-tuned on the task above $\mathcal{M_{\textsc{Rel}}}$, can it help in providing a better initialization for transfer learning an empirically superior model $\mathcal{M}'$ on other downstream tasks?
\begin{table}[t!]
    \scriptsize{\centering
        \begin{tabular}{m{10em}|m{3em}m{3em}m{3em}m{3em}}
            \toprule
            \textbf{Category} & \textbf{Datasets} & \textbf{Train} & \textbf{Val.} & \textbf{Test}\\
            \midrule
            \multirow{3}{*}{\textbf{Pairwise Symmetric}}&\textbf{QQP} & 327462 & 40430 & 36384 \\
            &\textbf{PAWS} & 49401 & 8000 & 8000 \\
            &\textbf{MRPC} & 3302 & 408 & 366 \\
            \midrule
            \textbf{Single Sentence} & \textbf{SST2} & 60615 & 6872 & 6734 \\
            \midrule
            \multirow{2}{*}{\textbf{Pairwise Non-symmetric}} &\textbf{QNLI} & 99506 & 5463 & 5237 \\
            &\textbf{RTE} & 2241 & 277 & 249 \\
            \bottomrule
        \end{tabular}
        \caption{\label{tbl:datasets} Datasets Statistics. Please refer to \refsec{sec:experiments}. }
    }
\end{table}

\subsection{Setup}
\label{subsec:setup}
\begin{figure}[h]
\centering
\includegraphics[width=0.46\textwidth]{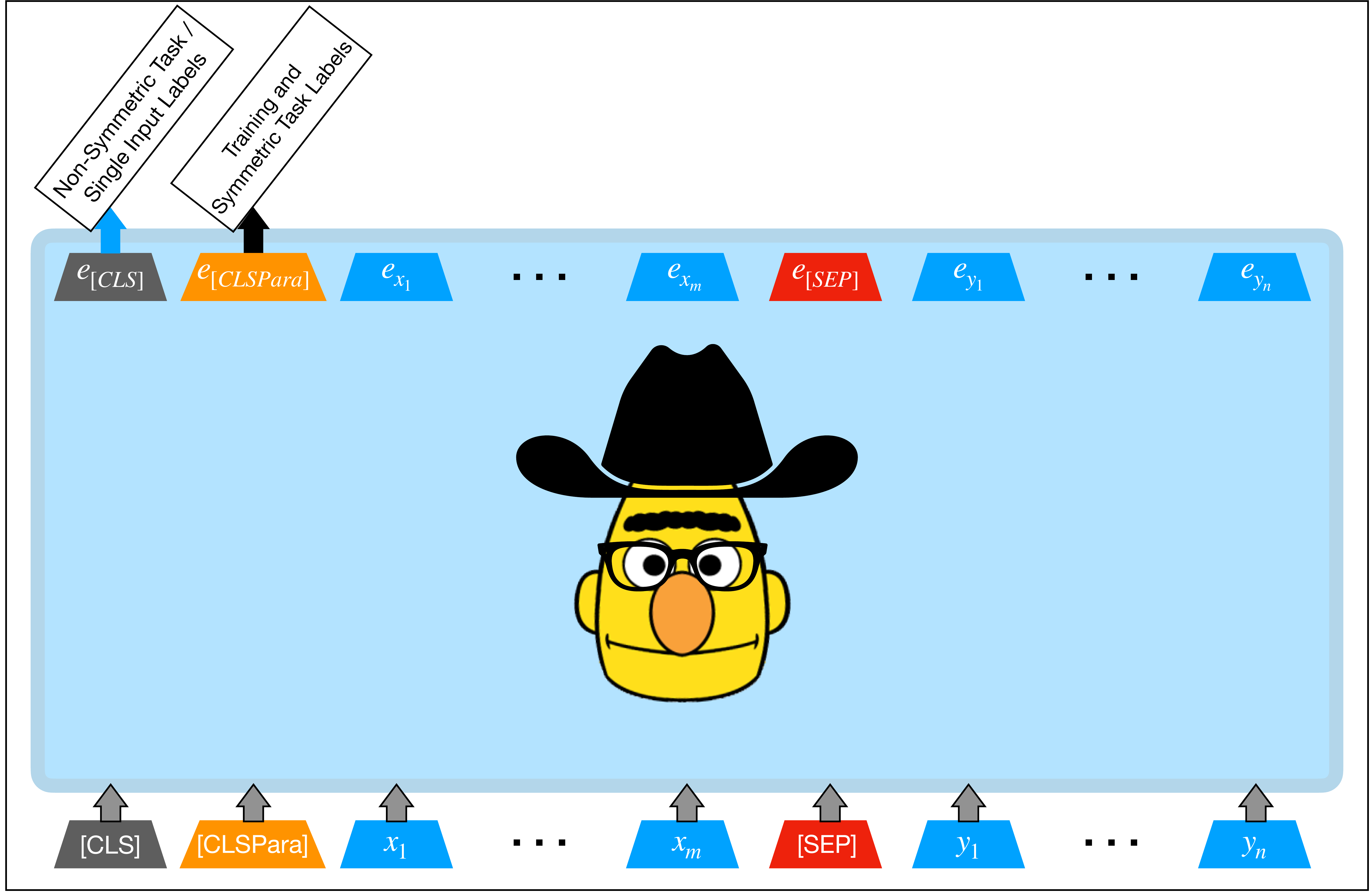}
\caption{BERT-with-consistency-loss. We use an additional classification token: \texttt{[CLSPara]} for our input, upon which the consistency objective is applied. Please refer \refsec{subsec:setup} for details.\label{fig:maindiag}}
\end{figure}
For problem \textbf{A} (\refsec{subsec:problem}), the input is a concatenation of tokenized strings $X = x_1, \ldots, x_m$ and $Y = y_1, \ldots, y_n$ separated using a special token (\texttt{[SEP]} in the case of BERT).  The concatenated inputs with the special token are passed through multiple self-attention layers \cite{vaswani2017attention}. In the traditional approach, the representation of the first token (\texttt{<s>} or \texttt{[CLS]}) is passed through a fully connected classifier layer (the same final representation is used irrespective of the arity of the task inputs). In our approach, we use the \texttt{[CLSPara]} representation for symmetric classification tasks whereas we use the standard first token (\texttt{<s>} or \texttt{[CLS]}) representation for single input and non-symmetric classification tasks (\refsec{sec:experiments}). Since we first fine-tune the model on \texttt{[CLSPara]} representation, our approach allows for pair-wise knowledge to be transferred to other downstream classification tasks (problem \textbf{B} (\refsec{subsec:problem})). 

We call this method BERT-with-consistency-loss and is shown in Figure~\ref{fig:maindiag}. Contrasting this with a traditional BERT-based approach, we see that, in the traditional BERT-based approach approach, the input is pre-pended with another special symbol (\texttt{[CLS]} in case of BERT and \texttt{<s>} in case of RoBERTa). In BERT-with-consistency-loss, we concatenate an extra symbol with the special symbol. We call the extra symbol \texttt{[CLSPara]}. This extra token is specifically used for symmetric classification tasks to ensure consistency of prediction.
\noindent The standard objective used for fine-tuning BERT-based models is the cross-entropy loss, which maximizes the probability of predicting the correct output class for a given input, given as:
\begin{equation}
    \mathcal{L}_{ce}(y, \hat{y}) = - \sum_i y_i \log \hat{y_i},
    \label{eqn:cross}
\end{equation}
where $y$ is the one-hot representation of the target class, $\hat{y}$ is the softmax output of the model, and $i$ is the associated co-ordinate. As described earlier, this objective may produce an inconsistent prediction based on the order of the two inputs. To overcome this weakness, we propose an additional consistency loss formulated in terms of either the KL or the JS Divergence. 
We pass the inputs $X$ and $Y$ through the same model twice, once as a pair $(X,Y)$ (called $L2R$) and then as the pair $ (Y,X)$ (called $R2L$).
Having obtained the outputs from the model for $L2R$ and $R2L$, the final objective function for \consistentbert{} is as follows:
\begin{equation}
\begin{split}
    \mathcal{L} &= \mathcal{L}_{ce}(y, \hat{y}_{L2R}) + \mathcal{L}_{ce}(y, \hat{y}_{R2L}) \\
    &+ \lambda*\mathcal{D}(p_{L2R}||p_{R2L}),
    \label{eqn:overall}
\end{split}    
\end{equation}

where $\lambda$ is the weight assigned to the consistency loss, $p_{L2R}$ and $p_{R2L}$ are the associated confidence/softmax vectors assigned by the model for $L2R$ and $R2L$ sentence pairs, and $\mathcal{D}$ is one of the following:
\begin{enumerate}
\setlength\itemsep{0.1em}
    \item $KL(p||q) = \sum_{x \in X} p(x) \log\frac{p(x)}{q(x)}$
    \item $JS(p||q) = \frac{1}{2}KL(p||m) + \frac{1}{2}KL(q||m)$,
    \label{eqn:js}
\end{enumerate}
Here $p, q$ are probability distributions and $m = \frac{1}{2}(p+q)$. Minimizing divergences between two distributions brings them closer to each other.

%% file: sections/experiment.tex
\section{Experimental Setup}
\label{sec:experiments}

\begin{table*}[ht!]
    \scriptsize{\centering
        \begin{tabular}{P{10.4em}P{7em}P{7em}P{7em}|P{7em}P{7em}P{7em}}
            \toprule
            &\multicolumn{3}{c|}{\textbf{(A) $L2R$ and $R2L$ Prediction Consistency }} & \multicolumn{3}{c}{\textbf{(B) $L2R$ and $R2L$ Confidence Consistency}} \\
            & \multicolumn{3}{c|} {Mean \textpm{} stddev (\refsec{subsec:evaluation}: Evaluation [1])}& \multicolumn{3}{c}{Pearson Correlation [MSE * 1000] (\refsec{subsec:evaluation}: Evaluation [2])} \\
            \midrule
            \textbf{Models} & \textbf{QQP} & \textbf{PAWS} & \textbf{MRPC} & \textbf{QQP} & \textbf{PAWS} & \textbf{MRPC} \\
            \midrule
      \textsc{Bert-base} & 96.6 \textpm{} 0.15 & 96.0 \textpm{} 0.54 & 91.1 \textpm{} 1.41 &  98.2 [5.89]  & 96.5 [14.2]  & 92.7 [17.0] \\
            \textsc{BERT-base w/ KL} & \textbf{99.3 \textpm{} 0.02} & \textbf{98.1 \textpm{} 0.12} & \textbf{97.7 \textpm{} 0.82}  & \underline{99.9 [0.12]} & \underline{99.6 [0.5]} & \underline{99.5 [0.3]} \\
            \textsc{BERT-base w/ JS} & \textbf{98.9 \textpm{} 0.05} & \textbf{98.1 \textpm{} 0.22} & \textbf{96.9 \textpm{} 0.93}  & \underline{99.8 [0.48]} & \underline{99.3 [1.9]} & \underline{99.0 [1.1]} \\
            \midrule
            \textsc{RoBERTa-base} & 97.0 \textpm{} 0.14 & 96.7 \textpm{} 0.25 & 91.5 \textpm{} 0.22 & 98.3 [5.90] & 97.4 [10.8] & 94.1 [16.3] \\
            \textsc{RoBERTa-base w/ KL}& \textbf{99.3 \textpm{} 0.03} & \textbf{98.9 \textpm{} 0.11} & \textbf{97.4 \textpm{} 0.78} & \underline{99.3 [0.10]} & \underline{99.7 [0.4]} & \underline{99.5 [0.3]}\\
            \textsc{RoBERTa-base w/ JS}& \textbf{99.1 \textpm{} 0.05} & \textbf{98.7 \textpm{} 0.23} & \textbf{96.7 \textpm{} 1.11} & \underline{99.8 [0.40]}  &\underline{99.6 [1.5]} &\underline{99.0 [1.3]} \\
            \midrule
            \midrule
            \multicolumn{7}{c}{\textbf{(C) Classification Performance Metrics} (\refsec{subsec:evaluation}: Evaluation [3])}\\
            \midrule
            \textbf{Models} & \textbf{QQP (Acc/F1)} & \textbf{PAWS (Acc/F1)} & \textbf{MRPC (Acc/F1)}  & \textbf{SST2 (Acc)} & \textbf{QNLI (Acc)} & \textbf{RTE (Acc)} \\
            \midrule
            \textsc{Bert-base} & 89.5 / 85.7 &  91.1 / 90.1 & 78.3 / 82.7 & 94.0 \textpm{} 0.10 & 87.9 \textpm{} 0.13  &  63.0 \textpm{} 1.33 \\
            \textsc{Bert-base w/ KL} & 87.1 / 82.3 & 88.0 / 86.8 & 73.0 / 80.7 & 94.1 \textpm{} 0.20 & 71.2 \textpm{} 4.15 & 51.6 \textpm{} 1.50  \\
            \textsc{Bert-base w/ JS} & 89.7 / 86.0 & 90.5 / 89.5 & 76.6 / 82.6 & 94.2 \textpm{} 0.42 & 74.5 \textpm{} 0.80 & 50.2 \textpm{} 16.90 \\
            \midrule
            \textsc{RoBERTa-base} & 90.2 / 87.2 & 92.6 / 91.7 & 82.4 / 86.0  & 94.4 \textpm{} 0.39 & 89.9 \textpm{} 0.47 & 70.6 \textpm{} 2.35 \\
            \textsc{RoBERTa-base w/ KL} & 87.2 / 82.7 & 91.5 / 90.5  & 74.7 / 81.0 & 94.5 \textpm{} 0.36 & 85.3 \textpm{} 1.62 & 58.7 \textpm{} 5.40 \\
            \textsc{RoBERTa-base w/ JS} & 90.0 / 86.6 & 92.3 / 91.6 & 79.2 / 84.9 & 95.1 \textpm{} 0.12 & 86.8 \textpm{} 1.51 & 61.4 \textpm{} 1.06  \\
            \bottomrule
        \end{tabular}
        \caption{\label{tbl:results}
        \textbf{Parts (A) \& (B)}: $L2R$ and $R2L$ Prediction and Confidence Consistency. \textbf{Part (C) Classification Metrics}. (*-BASE) indicate \bertsmall{}, (*- $W/$ *) indicate \consistentbertsmall{}.  Higher Accuracy, Higher Pearson Correlation and lower MSE are better. Numbers in \textbf{bold} are statistically significant. \underline{Underlined} numbers are better on average than baselines. Please refer to \refsec{subsec:quantitative} for a discussion.}
    }
\end{table*}

\subsection{Datasets}
\label{subsec:datasets}
We experiment with 5 standard datasets from the GLUE benchmark \cite{wang2019glue} as well as the PAWS dataset \cite{zhang2019paws}\footnote{Since the test split of these datasets is not available in the GLUE benchmark\cite{wang2019glue}, we use splits as given in \reftbl{tbl:datasets}. The validation dataset is kept as original and the new train and test sets are created by randomly splitting initial train data into train and test sets.}. We categorize them under the following headings:

\noindent \textbf{A. For Symmetric Tasks}: \textbf{(i) QQP:} Quora Question Pairs \cite{WinNT} data set contains pairs of questions marked with either 1 (paraphrases) or 0 (not paraphrases).\\ 
\textbf{(ii) PAWS:} Paraphrase Adv. from Word Scrambling \cite{zhang2019paws}, contains human labeled sentence pairs annotated in line with QQP. The uniqueness about this dataset is the creation procedure which involves back-translation and word swapping.
\textbf{(iii) MRPC:} Microsoft Research Paraphrase Corpus  \cite{dolan2005automatically} comprises human annotated sentence pairs collected from newswire articles.
\\ \textbf{B. For Single Input Task}: \textbf{(i) SST2:} Stanford Sentiment Treebank  \cite{socher2013recursive}. This is a collection of human-annotated movie reviews. We work with the standard two class setting where the annotations have opposite polarities (1 for positive sentiment and 0 otherwise).
\\ \textbf{C. For Non-symmetric tasks}: 
\textbf{(i) QNLI:}  Natural Language Inference dataset constructed from SQuAD \cite{rajpurkar2016squad} related to a two-class classification problem to determine if the premise entails a hypothesis or not.  \\
\textbf{(ii) RTE:} Recognizing Textual Entailment \cite{dagan2005pascal, bar2006second, giampiccolo2007third, bentivogli2009fifth} Corpus is a combination of multiple RTE datasets containing one of two labels (1  for entailment and 0 for  non-entailment).
\subsection{Evaluation}
\label{subsec:evaluation}
We analyse the results of the traditional objective as well as our approach on \textsc{BERT-Base} and \textsc{RoBERTa-Base} across four different seeds under the following categories:
\begin{enumerate}
\setlength{\itemsep}{0pt}
\item \textbf{Prediction Consistency:} This evaluation is done only for the symmetric task. $ \text{Score} = \frac{\mathbbm{1}_{(l_{L2R} = l_{R2L})}}{(\# \text{ of } L2R \text{ Samples})} * 100$, where $l_{L2R}, l_{R2L}$ denote labels for $L2R$ and $R2L$, respectively. Note that this is not related to the ground truth labels.
\item \textbf{Confidence Consistency:} We perform these evaluations specifically for symmetric task. This is to analyze how aligned are the confidence (softmax output associated with label 1) predicted by the model for $L2R$ and $R2L$ setting. The metrics used are the pearson correlation (scaled by 100) and the mean squared error (MSE - scaled by 1000) between the two confidence scores of the test data.
\item \textbf{Standard Classification Metrics:} These are task-specific metrics (accuracy/F1) used in the standard GLUE tasks \cite{wang2019glue}
\end{enumerate}

\subsection{Implementation Details}
\label{subsec:implementation}
To fine-tune the model for symmetric classification tasks, we club together three paraphrase detection datasets: (a) QQP, (b) PAWS, and (c) MRPC. To make sure that all the models see the same data, we augment the dataset with its reverse samples during training. The model is then trained by passing the \texttt{[CLSPara]} (\refsec{subsec:setup}) representation through a low-capacity classifier, and optimized using Equation \ref{eqn:cross} for baseline models and  Equation \ref{eqn:overall} for the consistency inducing models (Ours). We then use these models to conduct two sets of evaluations. We first evaluate the paraphrase detection results on QQP, PAWS, and MRPC individually. We then take the fine-tuned model obtained above and additionally fine-tune (\texttt{[CLS]} or \texttt{<s>} token) on the 
single input task (SST-2) and non-symmetric tasks (QNLI, RTE).

\noindent We use the hugging-face library \cite{wolf2020huggingfaces} for tokenizing the input, and the pytorch-lightning framework \cite{falcon2019pytorch} for loading the pre-trained models and fine-tuning them. We optimize the objective using the AdamW \cite{loshchilov2019decoupled} optimizer with a learning rate of 2e-5 (obtained through hyperparameter tuning \{2e-4, 2e-5, 4e-5, 2e-6\}). Since the input contains an additional token \texttt{[CLSPara]}, we extend the tokenizer vocabulary for each of the models. Each model was fine-tuned on a single Nvidia 1080Ti GPU (12 GB) for a maximum of 3 epochs ($\approx$ 6hrs/experiment). In case of BERT \cite{devlin2019bert}, we use the \texttt{bert-base-cased} model while for RoBERTa \cite{liu2019roberta}, we use the \texttt{RoBERTa-base} model.
For training stability, we perform lambda-annealing i.e., increase the $\lambda$ parameter from 0.0 to 100.0 as the training progresses. This ensures that the model has developed the capability to classify the sentence pairs with some degree of correctness before making it adhere to the appropriate symmetric confidence scores. We also experimented with fixed $\lambda$, but the resultant models were slow to converge ($\approx$ 15 epochs).

%% file: sections/results.tex
\section{Results}
\label{sec:results}
Our experiments address three questions: 
\begin{description}
\setlength{\itemsep}{0pt}
\item[Q1.] What are the shortcomings of the current objective function for symmetric classification tasks? (\refsec{sec:introduction}, \refsec{subsec:qualitative})
\item[Q2.] Does adding the consistency loss alleviate the inconsistency problem? (\refsec{subsec:quantitative})
\item[Q3.] Can consistency-based fine-tuning improve other downstream tasks? (\refsec{subsec:quantitative})
\end{description}

\subsection{Quantitative Analysis}
\label{subsec:quantitative}
\reftbl{tbl:results} presents our results. \textbf{Parts (A) \& (B)} compare $L2R$ and $R2L$ models in terms of prediction consistency and confidence consistency. Models trained with the consistency loss (indicated by $W/ *$) assign more similar predictions (indicated by higher scores in (A)) and confidence scores (indicated by higher correlation in (B)) as compared to the base model (indicated by $-BASE$), for both the base models (\textsc{BERT-base}/\textsc{RoBERTa-base}) and all symmetric test data sets (QQP, PAWS, MRPC). Moreover, the MSE (indicated within square brackets in part (B)) with consistency training is an order-of-magnitude smaller than without it.
The improvements in part (A) are  statistically significant at significance level ($\alpha$) of 0.01 according to McNemar's statistical test \cite{dror2018hitchhikers}.\\

\noindent \textbf{Part (C)} shows the results on  \textbf{downstream fine-tuning.} Our models (indicated by $W/ *$) do not compromise significantly (statistically evaluated) on the classification metrics for QQP, PAWS, and MRPC (F1/accuracy). The consistency loss does not change the accuracy scores of single sentence input tasks (SST-2), but affects the non-symmetric tasks (QNLI, RTE) negatively. This seems natural since the final objective of both the tasks is quite different and, in many cases, uncorrelated or negatively correlated. Incorporating consistency loss before fine-tuning on non-symmetric tasks (such as entailment) should, therefore, be avoided.\\
\noindent \textbf{Limitations:} Our goal is to increase the reliability (measured in terms of confidence scores) of the model and not specifically target classification performance metrics like accuracy and F1. Cases where they increase, can only partially be attributed to a stricter consistency constraint.

\subsection{Qualitative Analysis}
\label{subsec:qualitative}
We sample 30 instances that were assigned opposite labels for $L2R$ and $R2L$ by the \textsc{BERT-Base} models (majority voting) 
for QQP, MRPC and PAWS. An evaluator with NLP expertise analysed these examples and grouped them into recall error types. We then check the predictions for the same set of instances from \textsc{BERT + JS} (recall). Counts for these error types (defined in \refsec{subsec:errortypes}) are shown in Table~\ref{tbl:erroranal}. Out of those 30 examples for QQP, MRPC and PAWS, 26, 26 and 23 respectively get corrected by \consistentbertsmall{}. In general, the numbers reduce for all error types.
      
\begin{table}[t]
    \scriptsize{\centering
        \begin{tabular}{m{24.8em}P{1em}P{1em}}
            \toprule
        \bf Error type & \bf \bertsmall{} & \bf \consistentbertsmall{} \\  \midrule
        \multicolumn{3}{c}{\textbf{QQP}}\\ \midrule
        Different expected answer & 	4& 	0\\
        Different answer type + Additional details	& 8& 	1\\
        Different answer type + Additional details + Pronoun change	& 1& 	0\\
        Additional details and/or pronoun change & 	17 & 	3\\
       \midrule
        \multicolumn{3}{c}{\textbf{MRPC}}\\
        \midrule
        Additional details missing &	13 & 2\\
        Reordering of phrases	& 3 & 0\\
        Named entities and pronouns	 &6 & 1 \\
        Focus of sentences is different &	6 & 0\\
        Synonyms	& 2 & 1\\ \midrule
        \multicolumn{3}{c}{\textbf{PAWS}}\\
        \midrule
        Phrases are changed &	10 &	4\\
        Nouns/adjectives are changed &	12&	1\\
        Nouns/adjectives and phrases are changed &	4&	0\\
        Named entities are changed &	3&	1\\
        Names entities and nouns/adjectives are changed &	1&	1\\
        \bottomrule
        \end{tabular}
    \caption{\label{tbl:erroranal}Recall errors in QQP, MRPC \& PAWS: BERT (\bertsmall{}) and BERT with JS (\consistentbertsmall{}). Please refer to \refsec{subsec:qualitative}.}
}
\end{table}

%% file: sections/conclusion.tex
\section{Conclusion}
\label{sec:conclusion}
In this paper, we proposed an additional objective: consistency loss between $L2R$ and $R2L$ predictions so as to alleviate the problem of input order-sensitive inconsistency in the case of symmetric classification tasks. 
For three symmetric classification tasks,
our proposed solution, BERT-with-consistency-loss, results in an improved consistency in terms of Pearson's correlation and MSE. As expected, consistency loss results in a drop in the performance of non-symmetric classification tasks such as QNLI and RTE. Surprisingly, using KL divergence results in marginally higher \textit{consistency} than the JS counterpart. We leave this analysis for future work. Our qualitative analysis shows that all error types, including change in phrases or addition/deletion of details are reduced when the consistency loss is incorporated.

While consistency loss ensures that the predicted labels are the same even if the order of inputs is swapped, it can be adapted in the future to ensure expected outputs for anti-symmetric classification tasks (where $\mathbb{P}(L2R) = 1 - \mathbb{P}(R2L)$) like next and previous sentence prediction, where reordering the inputs must result in an opposite predicted label. In addition, the proposed method can be applied to evaluate paraphrasing models~\cite{kumar-etal-2019-submodular, kumar2020syntax} as well. In order to validate that paraphrasing models are indeed generating semantically similar outputs, BERT-with-consistency-loss can be used to either evaluate and filter out incorrect generations or be used as an objective to train learned metrics like BLEURT \cite{sellam2020bleurt}.

%% file: sections/ethics.tex
\section*{Ethical Considerations}
\label{sec:ethics}

The primary aim of this work is to highlight the inconsistency in labels and confidence scores of generated by standard pre-trained models for symmetric classification tasks. To mitigate the aforementioned inconsistency, we propose a loss function that incorporates divergence between outputs when the input order is swapped. We do not anticipate any additional ethical issues being introduced by our loss function as compared to the original standard pre-trained models, specifically BERT and RoBERTa. All the datasets used in our experiments are subset of the datasets from previously published papers, and to the best of our knowledge, do not have any attached privacy or ethical issues. That being said, further efforts should be made to study the inherent biases encoded in the pre-trained language models and the datasets.

%% file: sections/appendix.tex
\section{Appendix}
\label{sec:appendix}

\begin{table*}[ht]
    \small{\centering
        \begin{tabular}{p{4em}p{28em}P{4em}P{4em}P{4em}}
            \toprule
            \textbf{Dataset} & \textbf{Example pair} & \textbf{True label} & \textbf{$L2R$ Label} & \textbf{$R2L$ Label}\\ \midrule
            \multirow{7}{*}{MRPC} & \textbf{(1)} \textit{Shares in Wal-Mart closed at \$ 58.28 , up 16 cents , in Tuesday trading on the New York Stock Exchange.}  \textbf{(2)} \textit{Wal-Mart shares rose 16 cents to close at \$ 58.28 on the New York Stock Exchange.} & 1 & 0 & 1\\
            \cmidrule{2-5}
             &   \textbf{(1)} \textit{Darren Dopp , a Spitzer spokesman , declined to comment late Thursday.}	\textbf{(2)} \textit{John Heine , a spokesman for the commission in Washington , declined to comment on Mr. Spitzer 's criticism.} & 0 & 0 & 1 \\
            \midrule
            \multirow{4}{*}{QQP} & \textbf{(1)} \textit{How do I retrieve my deleted history from Google chrome?} \textbf{(2)} \textit{Can history be retrieved after deleting Google chrome?} & 1 & 0 & 1   \\
            \cmidrule{2-5}
            & \textbf{(1)} \textit{Is consciousness possible without self-awareness?} \textbf{(2)} \textit{Is self-awareness possible without consciousness?} & 0 & 1 & 0 \\ 
            \midrule
            \multirow{9}{*}{PAWS} & \textbf{(1)} \textit{This iteration is larger and has a smaller storage capacity than its previous versions.} \textbf{(2)} \textit{This iteration is smaller and has a greater storage capacity than its previous versions} & 0 & 0 & 1  \\ 
            \cmidrule{2-5}
            & \textbf{(1)} \textit{To get there , take Marine Drive west from the Lions Gate Bridge past Horseshoe Bay to Lighthouse Park and then continue on to 7100 Block Marine Drive.}	\textbf{(2)} \textit{To get there , take the Marine Drive from the Lions Gate Bridge to the west , past the Horseshoe Bay , Lighthouse Park and continue on to the 7100 Marine Drive block.} & 1 & 1 & 0 \\
            \bottomrule
        \end{tabular}
        \caption{\label{tbl:examples} Sample pairs which are classified differently by the fine-tuned model based on their input order in the standard classification setting in each of the paraphrase dataset. Please refer \refsec{sec:introduction}, \refsec{subsec:setup} for details.}
}
\end{table*}

\subsection{Recall Error Types in Qualitative Analysis}
\label{subsec:errortypes}
The qualitative analysis compares types of errors with and without consistency loss. The recall error types can be described as follows:\\

\noindent \textbf{A. QQP}:
\begin{enumerate}
\item \textbf{Different expected answer}: This error is said to occur in the case of QQP when the two input questions have a different expected answer. An example of such a pair is: `\textit{Is consciousness possible without self-awareness?}' and `\textit{Is self-awareness possible without consciousness?}'. The two questions are essentially complements of each other.
\item \textbf{Different answer type + Additional details}: This error is said to occur when one of the inputs is structured in a way that the answer would solicit additional details. For example, the input pair `\textit{How do I structure a big PHP project?}' and `\textit{How do I build a perfect PHP project?}' are similar - but nuances between `structuring' and `building' a project may result in different answers.
\item \textbf{Additional details and/or pronoun change}: The input pair `\textit{What are the best ways to get thick and wavy hair?}'	and `\textit{How can I get thick, wavy hair (as a guy)?}' is similar - although the latter uses the first-person proverb.
\end{enumerate}
\noindent \textbf{B. MRPC}:
\begin{enumerate}
    \item \textbf{Additional details missing}: One of the inputs contains information (\textit{i.e.}, details) that are not present in the other input. For example, `\textit{The caretaker, identified by church officials as Jorge Manzon, was believed to be among the nine missing - some of them children}' contains the number of missing persons that are not present in `\textit{The caretaker, identified by church officials as Jorge Monzon, was believed to be among the missing, who are presumed dead}'.
    \item \textbf{Reordering of phrases}: The two inputs contain the same information although the information may be represented using different phrasal structures. For example, `\textit{Shares in Wal-Mart closed at \$ 58.28 , up 16 cents , in Tuesday trading on the New York Stock Exchange.}' conveys the same information as 	`\textit{Wal-Mart shares rose 16 cents to close at \$ 58.28 on the New York Stock Exchange .}' The former uses passive voice while the latter uses `shares' as the main verb.
    \item \textbf{Named entities and pronouns}: One input replaces entities with pronouns, as in the case of `\textit{The bonds traded to below 60 percent of face value earlier this year}' and	`\textit{They traded down early this year to 60 percent of face value on fears Aquila may default .}'
    \item \textbf{Focus of sentences is different}: While information in one input is subsumed by the other, the latter might focus on a broader context. For example, `\textit{A power cut in New York in 1977 left 9 million people without electricity for up to 25 hours}' is implied in the sentence	`\textit{The outage resurrected memories of other massive power blackouts , including one in 1977 that left about 9 million people without electricity for 25 hours .}' However, the latter describes a resurrection of memories of the event in 1977.
    \item \textbf{Synonyms}: One or more words in an input may be replaced by its synonyms in the other input. For example, `\textit{In 2001 , the number of death row inmates nationally fell for the first time in a generation}' can be converted to `\textit{In 2001 , the number of people on death row dropped for the first time in a decade.}' by replacing the word `fell' with `dropped'.
    \end{enumerate}
    \noindent \textbf{C. PAWS}
    \begin{enumerate}
    \item \textbf{Nouns/adjectives are changed}: In the case of these errors, adjectives are replaced. An example pair is `\textit{This iteration is larger and has a smaller storage capacity than its previous versions}' and `\textit{This iteration is smaller and has a greater storage capacity than its previous versions .}'
    \item \textbf{Named entities are changed}: This refers to pairs where named entities (locations/people) are different. An example is the pair `\textit{When Mexico was within Los Angeles , Botello was chief of staff for Mexican General Ramirez y Sesma . His two brothers also married daughters of the general}' and `\textit{When Los Angeles was within Mexico , Botello was Chief of Staff of the Mexican General Ramirez y Sesma , his two brothers also married the general 's daughters .}'
    \end{enumerate}